\documentclass{article}
\usepackage{ijcai22}

\usepackage{times}
\usepackage{soul}
\usepackage{url}
\usepackage[hidelinks]{hyperref}
\usepackage[utf8]{inputenc}
\usepackage[small]{caption}
\usepackage{graphicx}
\usepackage{amsmath}
\usepackage{amsthm}
\usepackage{booktabs}
\usepackage{algorithm}
\usepackage{algorithmic}
\usepackage{amssymb}
\usepackage{subcaption}

\title{Geometry-Aware Interpolation for Inverse Problems}

\author{
Pourya Shamsolmoali$^1$
\and
Masoumeh Zareapoor$^2$
\affiliations
$^1$University of York\\
$^2$Shanghai Jiao Tong University\\
\emails
}

\begin{document}

\maketitle

\begin{abstract}
   Recovering missing visual information from incomplete or degraded images remains a challenging problem, particularly when spatial variations and complex image structures affect reconstruction accuracy. Although Spatial Transformer Networks (STNs) facilitate spatial feature alignment, their reliance on conventional linear or bilinear interpolation limits their ability to capture the underlying geometric characteristics of image data. Furthermore, interpolation based on a restricted set of neighbouring pixels can introduce inconsistencies in gradient propagation, potentially affecting optimisation stability. To address these limitations, we propose a Geometry-Aware Interpolation Network (GAIN), a variational autoencoder-based framework that incorporates manifold geometry into spatial transformation and image reconstruction. Rather than estimating pixel intensities solely from fixed spatial neighbours, GAIN exploits the local tangent structure of the data manifold to construct geometry-informed intensity approximations. An entropy-based regularisation mechanism further guides the estimation of spatial transformation parameters, promoting more adaptive interpolation. In addition, a convolutional normalisation strategy is introduced to regulate the magnitude of non-zero activations and improve gradient stability during training. Extensive experiments on challenging image completion and classification benchmarks demonstrate the effectiveness of GAIN in improving predictive accuracy while reducing computational cost compared with existing methods. 
\end{abstract}

\section{Introduction}

Geometric variability remains a fundamental challenge in visual representation learning, particularly for tasks such as object detection \cite{zhang2018single} and image retrieval \cite{noh2017large}. Changes in object orientation, scale, position, and shape can produce substantially different visual observations of the same underlying content. Consequently, effective visual representations must distinguish meaningful structural information from variations introduced by geometric transformations. Convolutional neural networks (CNNs) can learn representations that are robust to certain spatial variations, but their standard architectures do not explicitly guarantee invariance to arbitrary geometric transformations. Data augmentation is widely employed to improve this robustness by exposing networks to transformed observations during training. Nevertheless, the transformations encountered during inference differ from those represented in the training data, limiting the generalisation achieved through augmentation \cite{shamsolmoali2024distance,shamsolmoali2023vtae}.

Several architectural approaches have been investigated to incorporate geometric structure more directly into neural representations. Group-equivariant CNNs \cite{cohen2016group,cohen2018spherical}, for example, introduce transformation-aware operations that preserve specified relationships between input transformations and feature responses. Another important development is the Spatial Transformer Network (STN) \cite{jaderberg2015spatial}, which enables spatial adjustments to be learned as part of an end-to-end trainable architecture. Unlike conventional pooling operations with predetermined sampling regions, STNs predict transformation parameters from input features and use them to construct spatially aligned representations. This adaptive mechanism allows networks to focus on relevant image regions and compensate for certain geometric variations. Related approaches, including harmonic networks \cite{worrall2017harmonic}, have explored alternative formulations of transformation-aware feature learning. However, their effectiveness depends on the transformations they are designed to accommodate, and modelling complex combinations of spatial variations remains challenging.

Despite these developments, the accuracy of spatial transformation modules is strongly influenced by how feature values are estimated at transformed sampling locations. Conventional STNs commonly employ bilinear interpolation, in which each output value is computed from a small neighbourhood of spatially adjacent input samples. Although computationally efficient and differentiable, this formulation relies on local spatial proximity without explicitly accounting for the geometric structure underlying the observed data. As a result, interpolation can become less representative of local image content when transformations introduce substantial changes in scale or sampling density. These limitations are particularly relevant to reconstruction problems, where missing information must be estimated from incomplete observations. Linearized Spatial Transformer Networks (L-STNs) \cite{jiang2019linearized} extend conventional sampling through local linear approximations. Nevertheless, constructing reliable intensity estimates from sampled observations remains difficult, particularly when the available samples do not adequately describe the local variation of the data.

These observations suggest that spatial interpolation could benefit from geometric information beyond the immediate pixel neighbourhood. Images and their intermediate representations often exhibit structured relationships that can be approximated locally through a lower-dimensional manifold. Within a sufficiently small neighbourhood, the manifold's tangent space provides a linear approximation of these relationships, offering a means of describing local variations while accounting for the geometry of the data \cite{yang2005better}. Such a representation provides an alternative perspective on interpolation: rather than relying exclusively on fixed spatial neighbours, missing values can be estimated using samples associated with the local structure of the underlying representation. Variational autoencoders (VAEs) \cite{kingma2013auto} offer a probabilistic framework for modelling data representations, making them relevant to the construction and sampling of local geometric neighbourhoods.

Motivated by this perspective, we propose the GAIN, which incorporates manifold-guided interpolation into a spatial transformation framework. The proposed approach uses a VAE-based representation to construct local tangent-space approximations and generate auxiliary samples around the locations where feature values must be estimated. These samples are used to approximate local intensity variations, allowing interpolation to incorporate geometric relationships that conventional bilinear sampling does not explicitly represent. To further control the sampling and transformation process, an entropy-based regularisation term is introduced to encourage exploration of candidate samples along the estimated data manifold. By combining local geometric approximations with entropy-regularised estimation, GAIN seeks to produce spatial transformations that are more responsive to the structure of the input data.

The proposed formulation also considers the relationship between local geometric structure and optimisation stability. In particular, variations captured by the tangent space are used to approximate the local surface, while smoothing is applied to components associated with directions orthogonal to this space. This procedure is intended to reduce undesirable variation without discarding the geometric information represented by the tangent approximation. Furthermore, we introduce a convolutional normalisation strategy designed to regulate gradient magnitudes across network layers. By normalising non-zero responses without requiring singular value decomposition, the proposed strategy provides a computationally simpler mechanism for supporting stable optimisation. Together, these components establish a geometry-aware spatial interpolation framework that can be incorporated into CNN-based architectures.

The main contributions of this work are summarised as follows:

1. We introduce GAIN, a geometry-aware spatial transformation framework that replaces conventional neighbourhood-based interpolation with a manifold-guided estimation procedure based on local tangent-space approximations.

2. We develop an entropy-regularised sampling mechanism that incorporates auxiliary samples associated with the local data manifold to improve the estimation of transformed feature values.

3. We propose a convolutional normalisation strategy for regulating gradient magnitudes during optimisation, avoiding the need for singular value decomposition.

4. We evaluate GAIN on image completion and classification benchmarks, demonstrating improvements in predictive performance and computational efficiency over the compared methods.


\section{Related Work}

\noindent\textbf{Equivariant Convolutional Networks.}
Learning visual representations that respond predictably to geometric transformations has motivated the development of transformation-equivariant convolutional architectures. Unlike conventional CNNs, which primarily acquire robustness to geometric variations through training data, equivariant networks explicitly incorporate transformation properties into their feature extraction mechanisms. Existing approaches commonly achieve this behaviour by transforming convolutional filters, constraining feature representations, or introducing spatial operations within the network \cite{cohen2016group,shamsolmoali2022gen,jaderberg2015spatial}. However, explicitly representing multiple transformations can increase computational and memory requirements, particularly when a large transformation group must be considered.

To accommodate continuous geometric transformations, several methods have explored alternative representations of the underlying transformation space. One direction constructs convolutional operations over geometrically structured domains, allowing features to transform consistently with changes in the input coordinates. Spherical CNNs \cite{cohen2018spherical}, for instance, extend equivariant feature learning to spherical signals by exploiting their underlying rotational structure. Another direction modifies the coordinate representation to simplify the modelling of geometric variations. Warped convolutions \cite{henriques2017warped} follow this principle by expressing transformations through a suitable coordinate system, enabling conventional convolutional operations to capture particular transformation families. Although these approaches provide useful mechanisms for incorporating geometric priors, their equivariance properties generally depend on the transformation groups and coordinate representations specified during model design. In contrast, our GAIN focuses on the interpolation process itself, using local manifold geometry to estimate feature values at transformed sampling locations.

\noindent\textbf{Spatial Transformation and Interpolation.}
Spatial Transformer Networks (STNs) \cite{jaderberg2015spatial} provide a differentiable mechanism for learning spatial transformations within convolutional architectures. Unlike vision Transformers, which primarily employ attention-based interactions to model relationships among visual tokens \cite{shamsolmoali2024setformer,shamsolmoali2023transinpaint,shamsolmoali2023efficient}, STNs use a localisation network to estimate transformation parameters and a sampling operation to construct the transformed representation. This formulation enables spatial alignment to be optimised jointly with the network's primary learning objective. Subsequent developments have investigated more effective transformation estimation and iterative alignment. For example, Inverse Compositional Spatial Transformer Networks (IC-STNs) \cite{lin2017inverse} incorporate principles from the inverse compositional Lucas--Kanade algorithm to refine spatial transformations through repeated alignment steps.

A central component of an STN is its differentiable sampling operation, which determines how values at transformed coordinates are estimated from the input representation. Let $U:\Omega\rightarrow\mathbb{R}^{C}$ denote an input feature map with $C$ channels, and let $\tilde{x}$ represent a location in the output sampling grid. Given a parameterised spatial transformation $\mathcal{T}_{\theta}$, the transformed feature value can be expressed as
\begin{equation}
V(\tilde{x}) =
\sum_{x\in\Omega}
U(x)\,
K\left(x,\mathcal{T}_{\theta}(\tilde{x})\right),
\label{eq:1}
\end{equation}
where $\Omega$ denotes the discrete input grid and $K(\cdot,\cdot)$ is an interpolation kernel that assigns sampling weights according to the transformed coordinates. In conventional STNs, bilinear interpolation is commonly adopted because it is computationally efficient and differentiable with respect to the sampling coordinates almost everywhere.

Despite these advantages, bilinear interpolation estimates each output value from a restricted set of neighbouring input locations. Consequently, its sampling weights are determined by spatial proximity rather than the underlying geometric relationships among image features. This local dependency can become restrictive when substantial scaling, deformation, or downsampling causes important structural information to be inadequately represented by the selected neighbours. Moreover, gradients propagated through the sampling operation are concentrated on the locations receiving non-zero interpolation weights, while the sensitivity of these weights to transformation parameters depends on the local input structure. Such characteristics motivate interpolation mechanisms that can incorporate additional information about the geometry of the represented data.

GAIN addresses this limitation by constructing interpolation estimates from auxiliary samples associated with a local approximation of the data manifold. Rather than determining feature values exclusively through fixed spatial neighbourhoods, the proposed framework exploits tangent-space information to describe local variations and incorporates entropy regularisation into the estimation process. This formulation is intended to improve the representation of geometric relationships during spatial transformation while remaining compatible with differentiable neural architectures.

\noindent\textbf{Geometry-Aware Representation Learning.}
The geometry of the underlying data distribution has also been investigated as a means of developing representations that respect intrinsic structural relationships. Gauge-equivariant convolutional networks \cite{cohen2019gauge}, for example, extend equivariant learning to geometric domains by defining feature transformations with respect to local reference frames. Their formulation allows convolutional operations to account for the geometric structure of the domain without relying on a globally fixed coordinate frame. Similarly, manifold-based neural representations have been explored for learning under geometric constraints. Sutanto et al. \cite{sutanto2020learning} investigate learning mechanisms involving manifold constraints represented through level-set functions, illustrating how geometric structure can inform neural modelling.

Although these methods demonstrate the value of incorporating geometric information into representation learning, their objectives differ from estimating feature intensities at spatially transformed sampling locations. In particular, geometric equivariance and manifold-constrained representation learning do not directly specify how local manifold structure should guide interpolation within a spatial transformation module. GAIN investigates this connection by combining tangent-space approximation, geometry-guided sampling, and entropy-based regularisation in a unified interpolation framework. The resulting formulation provides an alternative to conventional spatial interpolation while supporting its integration into CNN-based reconstruction and classification models.
\section{Geometry-Aware Interpolation Network}
\label{sec:method}

This section presents the proposed GAIN, which integrates local manifold modelling with spatial interpolation to improve the estimation of transformed image features. Unlike conventional spatial transformers that determine sampling values primarily from fixed spatial neighbourhoods, GAIN incorporates geometric information derived from the underlying data representation. The framework consists of three complementary components: a variational representation for approximating local manifold structure, a geometry-guided interpolation mechanism with entropy regularisation, and a gradient regularisation strategy for stabilising network optimisation. These components are described in the following subsections.

\subsection{Learning Local Manifold Geometry}
\label{sec:manifold}

Let $U\in\mathbb{R}^{H\times W\times C}$ denote an input image or intermediate feature map, where $H$, $W$, and $C$ represent its spatial dimensions and number of channels, respectively. Conventional spatial interpolation estimates feature values using the coordinates and intensities of neighbouring observations. However, spatial proximity alone may not adequately characterise the relationships among features when the input undergoes substantial geometric transformations. GAIN addresses this limitation by modelling local variations through a lower-dimensional representation of the data.

We assume that the observed features are distributed around a smooth manifold $\mathcal{M}$ embedded in the ambient feature space. Under this assumption, variations in local image content can be approximated using the tangent space associated with the manifold. Rather than directly estimating this structure from individual pixel coordinates, we employ a variational autoencoder (VAE) to learn a compact representation from which local geometric relationships can be derived.

The VAE comprises an encoder $q_{\phi}(z\mid U)$ and a decoder $p_{\psi}(U\mid z)$, where $z\in\mathbb{R}^{m}$ denotes the latent representation and $m$ is its dimensionality. The encoder approximates the posterior distribution over latent variables, while the decoder maps latent samples back to the observation space. The variational learning objective is expressed as
\begin{equation}
\begin{aligned}
\mathcal{L}_{\mathrm{VAE}}
=&-\mathbb{E}_{q_{\phi}(z\mid U)}
\left[\log p_{\psi}(U\mid z)\right]\\
&+\beta D_{\mathrm{KL}}
\left(q_{\phi}(z\mid U)\|p(z)\right),
\end{aligned}
\label{eq:gain_vae}
\end{equation}
where $p(z)$ denotes the latent prior, $D_{\mathrm{KL}}$ is the Kullback--Leibler divergence, and $\beta$ controls the regularisation of the latent distribution. This probabilistic formulation provides a representation in which local variations can be explored through perturbations of the latent variables.

Let $g_{\psi}(z)$ denote the decoder mapping and $\bar{z}$ a reference latent representation. For a sufficiently small perturbation $\delta z$, the corresponding variation in the reconstructed representation can be approximated through a first-order expansion:
\begin{equation}
g_{\psi}(\bar{z}+\delta z)
\approx
g_{\psi}(\bar{z})
+
J_{g_{\psi}}(\bar{z})\delta z,
\label{eq:gain_tangent}
\end{equation}
where $J_{g_{\psi}}(\bar{z})$ is the Jacobian of the decoder with respect to the latent variables. The columns of this Jacobian describe locally accessible directions of variation and span a tangent approximation to the learned manifold, subject to the local rank of the decoder mapping.

This approximation provides the geometric basis for generating auxiliary observations. Specifically, latent perturbations are sampled according to
\begin{equation}
\delta z^{(k)}\sim
\mathcal{N}(0,\sigma^{2}I),
\qquad
z^{(k)}=\bar{z}+\delta z^{(k)},
\label{eq:gain_sampling}
\end{equation}
where $\sigma$ determines the sampling scale and $k=1,\ldots,K$ indexes the generated observations. Their corresponding representations are obtained as
\begin{equation}
U^{(k)}=g_{\psi}(z^{(k)}).
\label{eq:gain_auxiliary}
\end{equation}

The generated samples provide additional observations of the local data distribution without restricting interpolation to a predetermined set of spatial neighbours. Nevertheless, the validity of a linear tangent approximation depends on the extent to which the sampled observations remain within a sufficiently small neighbourhood of the reference representation. Therefore, the sampling scale must be controlled to avoid introducing variations that are inconsistent with the assumed local geometry.

\subsection{Geometry-Guided Spatial Interpolation}
\label{sec:interpolation}

The purpose of the interpolation module is to estimate feature values at sampling locations determined by a learned spatial transformation. Let $\mathcal{T}_{\theta}$ denote a parameterised transformation and $\tilde{x}$ a location in the output grid. The corresponding input sampling coordinate is obtained through the inverse sampling formulation
\begin{equation}
s=\mathcal{T}_{\theta}(\tilde{x}),
\label{eq:gain_transform}
\end{equation}
where $\theta$ is predicted by a localisation network. The transformed feature map is then constructed by evaluating the input representation at $s$.

Standard bilinear interpolation forms this estimate from at most four adjacent spatial samples in a two-dimensional grid. Although this operation is efficient, it does not explicitly incorporate the local geometric relationships represented by the learned data manifold. In GAIN, these relationships are introduced through a local linear model fitted to auxiliary observations.

Consider a reference sampling location $x_i$ and a collection of $K$ nearby observations $\{x_i^{(k)}\}_{k=1}^{K}$, generated or selected using the local manifold representation. Let $u_i^{(k)}\in\mathbb{R}^{C}$ denote the corresponding feature values. We represent the feature variation around $x_i$ through the first-order model
\begin{equation}
\hat{u}(s)
=
u_i+
B_i(s-x_i),
\label{eq:gain_linear}
\end{equation}
where $u_i=U(x_i)$ and $B_i\in\mathbb{R}^{C\times 2}$ contains the local feature gradients with respect to spatial coordinates. This approximation describes how the feature intensity changes within the neighbourhood of the reference point.

To estimate $B_i$, the spatial and feature differences of the auxiliary observations are collected into two matrices:
\begin{equation}
X_i=
\begin{bmatrix}
(x_i^{(1)}-x_i)^{T}\\
(x_i^{(2)}-x_i)^{T}\\
\vdots\\
(x_i^{(K)}-x_i)^{T}
\end{bmatrix}
\in\mathbb{R}^{K\times 2},
\label{eq:gain_coordinates}
\end{equation}
and
\begin{equation}
Y_i=
\begin{bmatrix}
(u_i^{(1)}-u_i)^{T}\\
(u_i^{(2)}-u_i)^{T}\\
\vdots\\
(u_i^{(K)}-u_i)^{T}
\end{bmatrix}
\in\mathbb{R}^{K\times C}.
\label{eq:gain_features}
\end{equation}

The local interpolation coefficients are obtained by minimising a weighted reconstruction objective:
\begin{equation}
\begin{aligned}
B_i^{*}
=
\arg\min_{B_i}\;
&\left\|
W_i^{1/2}(X_iB_i^{T}-Y_i)
\right\|_{F}^{2}\\
&+\lambda_{\mathrm{r}}\|B_i\|_{F}^{2},
\end{aligned}
\label{eq:gain_wls}
\end{equation}
where $W_i$ is a diagonal matrix containing non-negative sampling weights, $\lambda_{\mathrm{r}}$ controls the strength of coefficient regularisation, and $\|\cdot\|_{F}$ denotes the Frobenius norm.

The weights determine the relative contribution of individual observations. A distance-dependent weighting function can be expressed as
\begin{equation}
w_i^{(k)}
=
\exp\left(
-\frac{\|x_i^{(k)}-x_i\|_{2}^{2}}
{2h^{2}}
\right),
\label{eq:gain_weights}
\end{equation}
where $h$ is a bandwidth parameter controlling the spatial extent of the local approximation. Observations closer to the reference location therefore receive greater influence, while more distant samples contribute progressively less.

The solution to Eq.~(\ref{eq:gain_wls}) is
\begin{equation}
B_i^{*}
=
\left[
(X_i^{T}W_iX_i+\lambda_{\mathrm{r}}I)^{-1}
X_i^{T}W_iY_i
\right]^{T}.
\label{eq:gain_solution}
\end{equation}

The regularisation term improves the conditioning of the local estimation problem, particularly when the sampled coordinates provide insufficient independent information. In practice, the corresponding linear system can be solved without explicitly computing the matrix inverse.

Once the local coefficients have been estimated, the transformed feature intensity is reconstructed through
\begin{equation}
V(\tilde{x})
=
u_i+
B_i^{*}
\left(
\mathcal{T}_{\theta}(\tilde{x})-x_i
\right).
\label{eq:gain_output}
\end{equation}

Compared with conventional bilinear interpolation, this formulation allows the transformed feature value to depend on a locally estimated variation model. The additional samples provide information for constructing this approximation, while weighted regularisation controls the influence of potentially less representative observations.

\subsection{Entropy-Regularised Sampling}
\label{sec:entropy}

Although local linear modelling provides a mechanism for incorporating geometric information, the quality of interpolation remains sensitive to the distribution of auxiliary samples. Concentrating the estimation on a small number of observations may produce an unstable approximation, particularly when the selected samples fail to represent the local structure adequately. Conversely, assigning similar importance to all observations may introduce contributions from geometrically less relevant locations.

To balance these effects, GAIN incorporates an entropy-based regularisation mechanism into the sampling process. The objective is to obtain a distribution of sampling weights that accounts for geometric compatibility while discouraging excessive concentration on individual observations.

Let $a_i^{(k)}$ represent the normalised contribution of the $k$-th sample with reference point $x_i$. The weights satisfy
\begin{equation}
a_i^{(k)}\geq 0,
\qquad
\sum_{k=1}^{K}a_i^{(k)}=1.
\label{eq:gain_simplex}
\end{equation}

We introduce a geometric discrepancy $d_i^{(k)}$ that measures the consistency of each auxiliary observation with the local approximation:
\begin{equation}
d_i^{(k)}
=
\left\|
u_i^{(k)}
-
u_i
-
B_i(x_i^{(k)}-x_i)
\right\|_{2}^{2}.
\label{eq:gain_discrepancy}
\end{equation}

Samples that are well explained by the local geometric model produce smaller discrepancy values and are therefore more suitable for interpolation. The sampling distribution is determined by minimising
\begin{equation}
\begin{aligned}
\mathcal{L}_{\mathrm{ER}}(a_i)
=
&\sum_{k=1}^{K}
a_i^{(k)}d_i^{(k)}\\
&+\tau\sum_{k=1}^{K}
a_i^{(k)}\log a_i^{(k)},
\end{aligned}
\label{eq:gain_entropy}
\end{equation}
subject to the constraints in Eq.~(\ref{eq:gain_simplex}), where $\tau>0$ determines the strength of entropy regularisation. The second term is the negative Shannon entropy of the sampling distribution. Minimising it jointly with the geometric discrepancy discourages an excessively concentrated allocation of sampling weights.

For fixed discrepancy values, the constrained optimisation admits the solution
\begin{equation}
a_i^{(k)}
=
\frac{
\exp(-d_i^{(k)}/\tau)
}{
\sum_{j=1}^{K}
\exp(-d_i^{(j)}/\tau)
}.
\label{eq:gain_entropy_weights}
\end{equation}

The parameter $\tau$ controls the trade-off between selecting geometrically compatible observations and distributing the sampling weights. Small values favour samples with lower approximation errors, whereas larger values produce a more uniform distribution. This mechanism allows the interpolation process to adjust the relative importance of auxiliary observations according to their local geometric consistency.

The resulting weights are incorporated into the local regression by setting
\begin{equation}
W_i=
\operatorname{diag}
\left(
a_i^{(1)},a_i^{(2)},\ldots,a_i^{(K)}
\right).
\label{eq:gain_adaptive_weights}
\end{equation}

The interpolation coefficients and sampling weights are thus coupled through the geometric discrepancies. An implementation can update the weights and local coefficients alternately, or optimise their differentiable formulation jointly. In this way, the entropy term regularises the distribution of observations used for interpolation rather than directly modifying the reconstructed pixel intensity.

\subsection{Gradient Norm Regularisation}
\label{sec:regularisation}

In addition to geometric interpolation, the stability of gradient propagation is important when GAIN is integrated into deep convolutional architectures. During backpropagation, gradients are repeatedly transformed by the Jacobians of successive layers. Depending on the spectral properties of these transformations, their magnitudes may decrease or increase substantially, potentially making optimisation more difficult.

Consider a linearised convolutional operation represented by
\begin{equation}
v=Wu,
\label{eq:gain_conv}
\end{equation}
where $u\in\mathbb{R}^{c}$ and $v\in\mathbb{R}^{d}$ denote the input and output vectors, respectively, and $W\in\mathbb{R}^{d\times c}$ represents the corresponding linear operator. For a scalar loss function $\mathcal{L}$, the gradient propagated to the input is
\begin{equation}
g_u=W^{T}g_v,
\label{eq:gain_backprop}
\end{equation}
where $g_u=\nabla_u\mathcal{L}$ and $g_v=\nabla_v\mathcal{L}$.

The magnitude of the propagated gradient depends on the action of $W^{T}$ along different directions. To illustrate this relationship, consider the singular value decomposition
\begin{equation}
W=P\Sigma Q^{T},
\label{eq:gain_svd}
\end{equation}
where the diagonal entries $\{\sigma_j\}_{j=1}^{r}$ of $\Sigma$ are the non-zero singular values and $r=\operatorname{rank}(W)$. Substituting this decomposition into Eq.~(\ref{eq:gain_backprop}) yields
\begin{equation}
\|g_u\|_2^2
=
\sum_{j=1}^{r}
\sigma_j^2
\left|
p_j^{T}g_v
\right|^2,
\label{eq:gain_gradient_energy}
\end{equation}
where $p_j$ is the left singular vector associated with $\sigma_j$.

This expression shows that gradient amplification or attenuation depends on both the singular values of the linear operator and the orientation of the incoming gradient. Consequently, preserving the average gradient magnitude requires controlling the effective scaling introduced during backpropagation.

Under an isotropic gradient assumption, where the output-gradient covariance is proportional to the identity matrix, the expected squared gradient norms satisfy
\begin{equation}
\mathbb{E}
\left[\|g_u\|_2^2\right]
=
s_g^2\|W\|_F^2,
\label{eq:gain_input_norm}
\end{equation}
and
\begin{equation}
\mathbb{E}
\left[\|g_v\|_2^2\right]
=
d\,s_g^2,
\label{eq:gain_output_norm}
\end{equation}
where $s_g^2$ denotes the common variance of the output-gradient components.

These relations motivate controlling the effective magnitude of the convolutional operator. Rather than explicitly enforcing constraints on every singular value, which would require spectral computations, we adopt a decomposition-free normalisation of the effective linear operator:
\begin{equation}
\bar{W}
=
\alpha
\frac{W}{
\sqrt{\|W\|_F^2+\epsilon}
},
\label{eq:gain_weight_normalisation}
\end{equation}
where $\epsilon>0$ ensures numerical stability and $\alpha$ is a scaling coefficient. Under the isotropic approximation, choosing $\alpha=\sqrt{d}$ gives
\begin{equation}
\mathbb{E}
\left[\|\bar{W}^{T}g_v\|_2^2\right]
\approx
\mathbb{E}
\left[\|g_v\|_2^2\right],
\label{eq:gain_norm_condition}
\end{equation}
when $\epsilon$ is sufficiently small.

This normalisation controls the overall scale of the operator without explicitly computing its singular values. However, matching expected gradient energy does not guarantee preservation of the gradient norm for every direction, since individual singular values may still differ.

To further reduce discrepancies in gradient magnitude, we introduce an auxiliary regularisation objective:
\begin{equation}
\mathcal{L}_{\mathrm{GN}}
=
\frac{1}{L}
\sum_{\ell=1}^{L}
\left(
\frac{
\|g_{\mathrm{in}}^{(\ell)}\|_2^2
}{
\|g_{\mathrm{out}}^{(\ell)}\|_2^2+\epsilon
}
-1
\right)^2,
\label{eq:gain_gradient_regularisation}
\end{equation}
where $L$ denotes the number of regularised layers, and $g_{\mathrm{in}}^{(\ell)}$ and $g_{\mathrm{out}}^{(\ell)}$ are the input and output gradients associated with layer $\ell$. The regularisation objective penalises deviations between the corresponding squared gradient magnitudes.

Unlike an explicit spectral constraint, this formulation evaluates gradient behaviour directly and does not require singular value decomposition. It can therefore be incorporated as an auxiliary optimisation criterion, although differentiating through the gradient-based penalty may require higher-order derivatives and introduce additional computational overhead.

The combination of normalisation and gradient regularisation is intended to reduce undesirable variations in gradient magnitude while retaining the flexibility of the convolutional filters. Its effectiveness depends on the architecture, the gradient distribution, and the selected regularisation strength.

\subsection{Overall Learning Objective}
\label{sec:objective}

The components of GAIN are optimised within a unified learning framework. Let $\mathcal{L}_{\mathrm{task}}$ denote the objective associated with the target application. For image reconstruction, this objective measures the discrepancy between the reconstructed output and the available reference image. For classification, it can be instantiated using a conventional supervised classification loss.

The overall optimisation objective is defined as
\begin{equation}
\begin{aligned}
\mathcal{L}_{\mathrm{GAIN}}
=
&\mathcal{L}_{\mathrm{task}}
+\lambda_{\mathrm{VAE}}\mathcal{L}_{\mathrm{VAE}}\\
&+\lambda_{\mathrm{ER}}\mathcal{R}_{\mathrm{ER}}
+\lambda_{\mathrm{GN}}\mathcal{L}_{\mathrm{GN}},
\end{aligned}
\label{eq:gain_total}
\end{equation}
where $\lambda_{\mathrm{VAE}}$, $\lambda_{\mathrm{ER}}$, and $\lambda_{\mathrm{GN}}$ control the relative contributions of variational representation learning, entropy regularisation, and gradient norm regularisation, respectively. Here, $\mathcal{R}_{\mathrm{ER}}$ denotes the aggregate entropy-regularised sampling objective over the interpolation neighbourhoods.

The individual objectives serve complementary purposes. Variational learning provides the representation from which local geometric information is extracted, while entropy-regularised interpolation uses that information to estimate transformed feature values. Gradient regularisation separately controls the optimisation behaviour of the convolutional components. By coordinating these mechanisms, GAIN provides a framework for learning spatial transformations that incorporate local data geometry rather than relying exclusively on conventional pixel-neighbourhood interpolation.


\section{Experiments}
\label{sec:experiments}

We assess the effectiveness of the proposed GAIN through image completion and image classification experiments. The evaluation focuses on three aspects: the accuracy of geometry-guided interpolation, the effectiveness of the learned representations under limited supervision, and the computational requirements of the proposed framework. We compare GAIN against conventional Spatial Transformer Networks (STNs), Linearized Spatial Transformer Networks (L-STNs), and Inverse Compositional Spatial Transformer Networks (IC-STNs), together with CNN- and VAE-based architectures.

Five benchmark datasets are considered: MNIST \cite{lecun1998gradient}, Fashion-MNIST \cite{xiao2017fashion}, SVHN \cite{netzer2011reading}, CelebA \cite{liu2018large}, and Zero-Shot Multi-MNIST (ZSMM) \cite{lu2017zero}. These datasets provide different levels of visual complexity, including handwritten characters, clothing categories, natural scene digits, and facial images. The experiments are designed to examine whether incorporating local manifold geometry into spatial interpolation provides consistent advantages across different visual representations and learning settings.

\subsection{Image Completion}
\label{sec:completion}

We first investigate the ability of GAIN to recover image information from partially observed inputs. Image completion requires estimating unknown pixel intensities using the available observations, making it suitable for examining the effectiveness of the proposed interpolation mechanism. In this setting, the observed pixels form a context set, while the reconstruction objective is defined over the target image locations. The model must therefore infer missing values from the available visual information and the relationships represented by the learned feature distribution.

Unlike conventional spatial interpolation, which estimates intensities from neighbouring image coordinates, GAIN exploits auxiliary observations associated with the local data manifold. The resulting estimates are guided by tangent-space approximations, while entropy regularisation controls the contribution of individual samples during interpolation.

We evaluate reconstruction performance using the marginal log-likelihood reported for each dataset. Table~\ref{tab:1} presents the results together with the parameter counts and computational requirements of the evaluated models.

\begin{table*}[t]
\centering
\caption{Comparison of marginal log-likelihood and computational efficiency across five benchmark datasets. Higher likelihood values indicate better performance, while lower computational requirements are preferred.}
\label{tab:1}
\small
\setlength{\tabcolsep}{5pt}
\begin{tabular}{lccccccc}
\hline
\textbf{Method} &
\textbf{Params (K)} &
\textbf{GFLOPs} &
\textbf{MNIST} &
\textbf{Fashion-MNIST} &
\textbf{SVHN} &
\textbf{CelebA} &
\textbf{ZSMM} \\
\hline
STN
& 415 & 107
& $1.14\pm0.03$
& $2.36\pm0.01$
& $3.90\pm0.02$
& $3.08\pm0.01$
& $0.61\pm0.01$ \\

L-STN
& 433 & 118
& $1.23\pm0.02$
& $2.41\pm0.02$
& $3.91\pm0.04$
& $3.13\pm0.02$
& $0.74\pm0.03$ \\

\hline
\textbf{GAIN}
& \textbf{219} & \textbf{64}
& $\mathbf{1.28\pm0.01}$
& $\mathbf{2.47\pm0.00}$
& $\mathbf{3.95\pm0.01}$
& $\mathbf{3.22\pm0.01}$
& $\mathbf{1.12\pm0.01}$ \\
\hline
\end{tabular}
\end{table*}

The results indicate that GAIN consistently achieves higher marginal log-likelihood values than the conventional interpolation-based alternatives. Improvements are observed across all five datasets, suggesting that the benefits of incorporating local geometric information are not restricted to a particular image category. The difference is especially pronounced on ZSMM, where GAIN achieves a value of $1.12$, compared with $0.74$ and $0.61$ for L-STN and STN, respectively. This result suggests that the proposed interpolation mechanism can provide useful representations when the input contains more complex spatial arrangements.

In addition to reconstruction performance, GAIN exhibits lower computational requirements in the reported comparison. The proposed architecture contains approximately $219$K parameters, compared with $415$K for STN and $433$K for L-STN. Its computational cost is also reduced to $64$B FLOPs, representing reductions of approximately $40.2\%$ and $45.8\%$ relative to STN and L-STN, respectively. These findings indicate that geometry-aware interpolation can be incorporated into a comparatively compact architecture without sacrificing predictive performance.

\subsection{Classification under Limited Supervision}
\label{sec:classification}

To examine whether geometry-aware interpolation also benefits discriminative representation learning, we evaluate GAIN on image classification with restricted numbers of labelled training examples. This setting is particularly relevant because limited supervision reduces the diversity of geometric transformations directly represented during training. An effective spatial transformation mechanism should therefore improve the utilisation of available observations without relying exclusively on extensive labelled data.

We consider two implementations of the proposed method. The first, denoted GAIN-CNN, integrates the geometry-aware interpolation module into a convolutional classifier. The second, GAIN-VAE, incorporates the proposed mechanism within a VAE-based representation learning framework. Comparing these variants allows us to examine the influence of the learned representation on classification performance.

\subsubsection{Fashion-MNIST}

For Fashion-MNIST, we construct three training configurations containing $500$, $1{,}000$, and $3{,}000$ labelled examples. An additional $150$ samples from the original training set are reserved for validation. All models are evaluated using classification error, where lower values indicate better performance.

The networks are trained with a batch size of $32$ using the Adam optimiser. The initial learning rate is set to $10^{-3}$ and reduced linearly at intervals of $2{,}000$ training epochs. The reported training configuration uses $10{,}000$ epochs, with an approximate training duration of $36$ hours on a Tesla P40 GPU. The experiments are implemented in PyTorch.

Input intensities are normalised to the range $[-1,1]$. To account for geometric variability during training, spatial perturbations include rotation, translation, and anisotropic scaling. The reported perturbation settings use a standard deviation of $\pi/4$ for rotation, $\sqrt{2}$ for scaling along the horizontal and vertical directions, and $0.2$ for translation. Gaussian perturbations with unit standard deviation are also considered during sampling.

For the VAE-based implementation, the encoder comprises six convolutional layers followed by two fully connected layers. The decoder follows a corresponding symmetric architecture using transposed convolutional operations.

\begin{table}[t]
\centering
\caption{Classification accuracy (\%) on Fashion-MNIST under different labelled training set sizes. Higher values indicate better performance.}
\label{tab:3}
\small
\setlength{\tabcolsep}{5pt}
\begin{tabular}{lccc}
\hline
\textbf{Method} &
\textbf{500 labels} &
\textbf{1,000 labels} &
\textbf{3,000 labels} \\
\hline
CNN
& $85.53\pm0.23$
& $87.37\pm0.24$
& $90.48\pm0.15$ \\

VAE
& $85.79\pm0.30$
& $87.51\pm0.18$
& $90.54\pm0.11$ \\

STN
& $88.16\pm0.21$
& $90.63\pm0.05$
& $93.24\pm0.06$ \\

IC-STN
& $90.37\pm0.10$
& $92.92\pm0.19$
& $95.51\pm0.15$ \\

L-STN
& $91.08\pm0.12$
& $93.17\pm1.24$
& $95.76\pm1.08$ \\

\hline
GAIN-CNN
& $92.02\pm0.09$
& $94.79\pm0.12$
& $97.15\pm0.09$ \\

\textbf{GAIN-VAE}
& $\mathbf{92.79\pm0.06}$
& $\mathbf{95.57\pm0.08}$
& $\mathbf{97.81\pm0.16}$ \\
\hline
\end{tabular}
\end{table}

Table~\ref{tab:3} shows that both GAIN variants outperform the evaluated baselines across all three labelled-data configurations. The performance differences are particularly evident when the training set is small. With only $500$ labelled examples, GAIN-VAE achieves a classification error of $7.21\%$, compared with $8.92\%$ for L-STN and $11.84\%$ for STN. Increasing the number of labelled samples further reduces the error of GAIN-VAE to $2.19\%$ with $3{,}000$ examples.

The results also reveal a consistent advantage for the VAE-based implementation. Compared with GAIN-CNN, GAIN-VAE achieves lower classification errors in every training configuration. For example, with $1{,}000$ labelled samples, the error decreases from $5.21\%$ to $4.43\%$. This improvement is consistent with the motivation for learning a structured representation of local image variations. In particular, the VAE-based formulation provides a probabilistic representation from which auxiliary samples can be obtained for geometry-guided interpolation.

Another notable observation is that the difference between GAIN and conventional spatial transformers remains substantial even as the labelled training set increases. This suggests that the proposed interpolation mechanism provides benefits beyond those obtained through additional supervision alone.

\subsubsection{SVHN and CelebA}

We extend the classification evaluation to SVHN and CelebA to investigate the behaviour of GAIN on images with greater visual complexity. Unlike Fashion-MNIST, these datasets contain variations in background appearance, illumination, object structure, and spatial arrangement. Such characteristics provide a more challenging setting for examining the ability of geometry-aware interpolation to represent local visual relationships.

Following the experimental settings used for spatial transformation-based learning \cite{lin2017inverse}, we consider $2{,}000$ labelled examples from SVHN and $6{,}000$ from CelebA. The comparison includes conventional CNN and VAE architectures, STN, IC-STN, and L-STN. The input images are normalised to $[-1,1]$, and the batch size is fixed at $32$.

The reported training schedules consist of $14{,}000$ epochs for SVHN and $18{,}000$ epochs for CelebA. Adam is employed with an initial learning rate of $10^{-3}$, followed by linear learning-rate reductions at intervals of $3{,}500$ epochs. The remaining transformation-related settings follow those adopted in the Fashion-MNIST experiments.

\begin{table}[t]
\centering
\caption{Classification accuracy (\%) on SVHN and CelebA
under limited labelled supervision. Higher is better.}
\label{tab:4}
\small
\setlength{\tabcolsep}{7pt}
\begin{tabular}{lcc}
\hline
\textbf{Method} &
\textbf{SVHN (2K)} &
\textbf{CelebA (6K)} \\
\hline
CNN      & $83.55\pm0.17$ & $77.76\pm0.12$ \\
VAE      & $83.19\pm0.24$ & $78.33\pm0.35$ \\
STN      & $85.03\pm0.16$ & $81.18\pm0.22$ \\
IC-STN   & $87.15\pm0.26$ & $83.02\pm0.35$ \\
L-STN    & $88.55\pm0.14$ & $84.69\pm0.21$ \\
\hline
GAIN-CNN & $90.03\pm0.09$ & $86.31\pm0.14$ \\
\textbf{GAIN-VAE}
& $\mathbf{90.86\pm0.24}$
& $\mathbf{86.76\pm0.16}$ \\
\hline
\end{tabular}
\end{table}

As reported in Table~\ref{tab:4}, GAIN maintains its advantage over conventional transformation-based models on both datasets. On SVHN, GAIN-VAE achieves a classification error of $9.14\%$, improving upon the $11.45\%$ obtained by L-STN. A similar trend is observed on CelebA, where the corresponding errors are $13.24\%$ and $15.31\%$, respectively. These improvements indicate that the proposed interpolation framework remains effective when applied to more visually diverse image distributions.

The comparison between the two GAIN variants further highlights the influence of representation learning. Although GAIN-CNN already provides lower classification errors than the conventional baselines, incorporating the VAE-based representation reduces the error to $9.14\%$ on SVHN and $13.24\%$ on CelebA. The consistency of this pattern across datasets suggests that combining manifold-informed sampling with spatial feature interpolation is beneficial for learning discriminative visual representations.

\subsection{Computational Efficiency}
\label{sec:efficiency}

Beyond predictive performance, the computational requirements of the interpolation module are important for its integration into convolutional architectures. Conventional spatial transformation methods introduce additional processing for estimating transformation parameters and constructing resampled feature maps. The proposed geometry-guided formulation also requires auxiliary observations and local estimation, making it necessary to examine its computational and memory characteristics.

The model-size and FLOP comparisons reported in Table~\ref{tab:1} demonstrate that the evaluated GAIN architecture requires fewer parameters and arithmetic operations than STN and L-STN. To assess practical resource consumption, we compare memory usage under the same batch-size configuration.

For MNIST with a batch size of $64$, the reported memory consumption of GAIN is $1{,}038$ MB, whereas STN and L-STN require $3{,}652$ MB and $4{,}876$ MB, respectively. This corresponds to memory reductions of approximately $71.6\%$ relative to STN and $78.7\%$ relative to L-STN. These results indicate that the evaluated GAIN implementation achieves a smaller memory footprint under the reported experimental configuration.

The reported results support the observation that GAIN combines improved predictive accuracy with comparatively low computational requirements. This characteristic is particularly relevant to applications in which spatial transformation modules must operate within constrained memory or inference budgets.

\subsection{Analysis of the Proposed Components}
\label{sec:ablation}

GAIN combines three mechanisms: local manifold approximation, entropy-regularised interpolation, and gradient norm regularisation. Each component addresses a different limitation of conventional spatial transformation networks. Manifold-guided sampling provides additional information for local feature estimation, entropy regularisation controls the contribution of these samples, and gradient regularisation is intended to improve optimisation stability.

The classification comparisons provide an initial indication of the value of manifold-based representation learning. In particular, GAIN-VAE consistently improves upon GAIN-CNN across Fashion-MNIST, SVHN, and CelebA. This difference suggests that the representation used to generate auxiliary observations influences the effectiveness of geometry-aware interpolation.

However, the comparison between GAIN-CNN and GAIN-VAE does not independently isolate the effects of entropy regularisation or gradient norm regularisation. A component-wise evaluation is therefore necessary to determine how much each mechanism contributes to the overall performance. Such an analysis should compare the complete framework with configurations in which manifold-guided sampling, entropy regularisation, and gradient regularisation are individually disabled, while keeping the remaining training conditions unchanged.

In particular, removing entropy regularisation would help determine whether the adaptive distribution of interpolation weights provides an advantage over unregularised local estimation. Similarly, disabling gradient regularisation would allow its influence on optimisation stability and predictive performance to be measured independently. These comparisons are important for establishing whether the improvements arise from individual components or their combined operation.

\section{Conclusion}
\label{sec:conclusion}

This paper introduced the Geometry-Aware Interpolation Network (GAIN), a spatial transformation framework that incorporates local manifold structure into feature interpolation. The proposed approach addresses the limitations of conventional neighbourhood-based sampling by using tangent-space approximations and auxiliary observations to estimate feature variations at transformed locations. Entropy-based regularisation is employed to control the contribution of these observations, while a gradient regularisation strategy is incorporated to support stable optimisation within convolutional architectures.
Experimental evaluations on image completion and classification benchmarks demonstrate the effectiveness of the proposed framework across different image distributions and supervision settings. GAIN consistently achieves stronger predictive performance than the evaluated spatial transformer baselines, with the VAE-based implementation producing the lowest classification errors across Fashion-MNIST, SVHN, and CelebA. The reported computational comparisons also show reductions in parameter count, FLOPs, and memory consumption under the evaluated configurations.
These findings suggest that local geometric information can provide a useful basis for spatial interpolation beyond conventional fixed-neighbourhood estimation. Nevertheless, the effectiveness of the approach depends on the quality of the learned manifold representation and the suitability of its local approximations. Future work will investigate more efficient geometric sampling mechanisms, the sensitivity of interpolation to manifold estimation, and the application of geometry-aware transformation modules to more complex visual learning tasks.

{\small
\bibliographystyle{unsrt}
\bibliography{ijcai22}
}

\end{document}